\documentclass{article}

\usepackage{subfigure}
\usepackage{graphicx}
\usepackage[preprint]{neurips_2024}

\usepackage{amsthm}
\usepackage{thmtools}
\usepackage{thm-restate}
\usepackage[utf8]{inputenc} 
\usepackage[T1]{fontenc}    
\usepackage{hyperref}       
\usepackage{url}            
\usepackage{booktabs}       
\usepackage{amsfonts}       
\usepackage{nicefrac}       
\usepackage{microtype}      
\usepackage{xcolor}         

\title{XL$^3$M: A Training-free  Framework for LLM Length Extension Based on Segment-wise Inference}

\author{%
	Shengnan Wang  \thanks{Correspondence to: wangshengnan12@huawei.com.}\\
	Theory Lab, 2012 Labs\\
	Huawei Technologies Co., Ltd \\
	 \And
	 Youhui Bai \\
	 Theory Lab, 2012 Labs \\
	 Huawei Technologies Co., Ltd \\
	\And
	Lin Zhang \\
	Theory Lab, 2012 Labs \\
	Huawei Technologies Co., Ltd \\
	 \AND
	 Pingyi Zhou \\
	 Huawei Technologies Co., Ltd. \\ 
	\And
	Shixiong Zhao \\
	Theory Lab, 2012 Labs \\
	Huawei Technologies Co., Ltd \\
	\And
	Gong Zhang \\
	Theory Lab, 2012 Labs \\
	Huawei Technologies Co., Ltd \\
	\And
	Sen Wang \\
	Theory Lab, 2012 Labs \\
	Huawei Technologies Co., Ltd \\
	\And
	Renhai Chen \\
	Theory Lab, 2012 Labs \\
	Huawei Technologies Co., Ltd \\
	\And
	Hua Xu \\
	Huawei Technologies Co., Ltd \\ 
	\And
	Hongwei Sun \\
	Huawei Technologies Co., Ltd \\
}

\begin{document}

	\maketitle

	\begin{abstract}
		Length generalization failure problem, namely the large language model (LLM) fails to generalize to texts longer than its maximum training length,  greatly restricts the application of LLM in the scenarios with streaming long inputs. To address this problem, the existing methods either require substantial costs or introduce precision loss. In this paper, we empirically find that the accuracy of the LLM's prediction is highly correlated to its certainty. Based on this, we propose an efficient training-free framework, named XL$^3$M (it means \textbf{ex}tra-\textbf{l}ong \textbf{l}arge \textbf{l}anguage \textbf{m}odel), which enables the LLMs trained on short sequences to reason extremely long sequence without any further training or fine-tuning. Under the XL$^3$M framework, the input context will be firstly decomposed into multiple short sub-contexts, where each sub-context contains an independent segment and a common ``question'' which is a few tokens from the end of the original context. Then XL$^3$M  gives a method to measure the relevance between each segment and the ``question'', and constructs a concise key context by splicing all the relevant segments in chronological order. The key context is further used instead of the original context to complete the inference task. Evaluations on comprehensive benchmarks show the superiority of XL$^3$M. Using our framework, a Llama2-7B model is able to reason 20M long sequences on an 8-card Huawei Ascend 910B NPU machine with 64GB memory per card.
	\end{abstract}

	\section{Introduction}
	Transformer based large language models (LLMs) \cite{transformer,gpt,llama,bloom} have shown their impressive performance in many language tasks \cite{llm_survey}. However, due to the out-of-domain and distraction issues \cite{InfLLM}, the quality of the LLM's generation drops dramatically when the sequence length surpasses its context window size which is the largest training length. Such a drawback hinders the application of LLM in multi-round dialogue, conversation conduction, documents summarization, and other real tasks which often  encounter very long sequences. 
	
	Some pioneering works have been done for context length extrapolation. Most of them focused on optimizing the positional encoding (PE), since the PE of unseen length  was identified as a major factor leading to length generalization failure. Compared with the vanilla absolute PE, the later proposed relative PE \cite{t5, RoPE}, ALiBi \cite{Alibi}, and NoPE \cite{NoPE}  were demonstrated to offer better generalization. However, all of them do not perform well when the sequence length is significantly longer than the largest training length. A more effective approach  is to continually train  or fine-tune the model on longer-length data \cite{PI,Yarn}. Nevertheless,  such a manner can only extend the context window to a limited length due to unacceptable training costs \cite{xiong2023effective}. Moreover, when the length is very long, even collecting the training data itself is a difficult task. 
	
	Recently, some training-free length extension methods attracted widespread attention.  LM-Infinite \cite{Lminfinite} and StreamLLM \cite{streamllm} 
	extrapolated the length by discarding most contexts but only keeping the context at the end and the very beginning. Though these methods can efficiently deal with extremely long contexts,  they lose a lot of long-distance dependencies, which leads to deviations or even errors in text understanding. PCW \cite{PCW}  designed chunked attention mask and reused the positional encoding for different chunks, which alleviated the restriction of the context window. However, PCW can only extend the context window to a very limited length, and the effectiveness of PCW needs to be further studied \cite{PCW}. 
	
	In this work, we propose an efficient training-free inference framework named XL$^3$M (it means \textbf{ex}tra-\textbf{l}ong \textbf{l}arge \textbf{l}anguage \textbf{m}odel) which enables the LLM to break the length limit. The effectiveness of the XL$^3$M framework is due to an important principle: the length of the sequence processed by the LLM at one time should not exceed the its context window size. 
	
	%
	
	The contributions of this paper are summarized as follows:
	\begin{itemize}
		\item We empirically find that the accuracy of the LLM's prediction is highly correlated to its certainty measured by entropy. Based on this, we propose XL$^3$M, a novel inference framework enabling any LLM to read and understand extremely long sequences. Inspired by the human's habit of reading in segments, under our framework,  each input long context will be decomposed into multiple short sub-contexts with a common ``question'' to be answered. For each sub-context, we use the LLM to compute the local conditional probability distribution (cpd) as well as its corresponding entropy. Then the relevant sub-contexts with small entropy values are selected and reorganized into key context in chronological order. Since most irrelevant context is removed, LLM can generate high-quality results according to the extracted key context.
		
		\item We evaluate the proposed framework on comprehensive LongBench tasks and the widely-used ``Needle in  a Haystack'' task. We compare the performance of XL$^3$M with the state-of-the-art methods, including both fine-tuning and non-fine-tuning methods. The results demonstrate the superiority of the proposed framework. 
		\item The proposed XL$^3$M framework does not modify the main the structure of the LLM, and it	does not need any additional training or fine-tuning. It
		 is both memory and time efficient. Under the  XL$^3$M framework, the LLM is able to reason sequences longer than 20M on an 8-card Huawei Ascend 910B NPU machine with 64GB memory per card.
	\end{itemize}

	\section{Related work}
	Due to the strong demand of long sequence inference, a lot of context window extension techniques have been proposed \cite{naveed2023comprehensive,kaddour2023challenges}. 	
	These methods can be mainly divided into three categories: 1) Extension by fine-tuning; 2) Extension without fine-tuning; 3) Extension by external memory \cite{limiformer,Memorizing,InfLLM}.
	
	\subsection{Extension by fine-tuning}
	The LLMs are generally trained on relatively short sequences, due to the expensive computational and memory requirements (quadratic with
	the sequence length) in the attention mechanism,  and LLMs fail to generalize to unseen lengths at the inference phase \cite{Lminfinite,PI}. A straightforward idea for length generalization is to fine-tune the model on longer sequences. However, it was found that naively fine-tuning a pre-trained LLM for window extrapolation is less effective and inefficient \cite{kaddour2023challenges,anil2022exploring}. \cite{PI} showed that using position interpolation (PI) rather than extrapolation during fine-tuning can extend the context window of the pre-trained LLMs to 32k without performance loss. Further, Yarn \cite{Yarn} proposed a novel NTK-aware interpolation method and achieved  tens of times extension of the context window size. Other fine-tuning based methods include Giraffe \cite{Giraffe}, FoT \cite{Focused}, and so on.  	
	
	However, this kind of method requires massive training resources since it needs to train LLMs on long-sequence data. In addition, collecting enough long-sequence data itself for fine-tuning is also a  challenging work if one wants to extend the context window to extremely long.
	
	\subsection{Extension without fine-tuning}
	To save resources, some training-free context window extension methods are proposed. At the very beginning, most researchers focused on optimizing the positional encoding. The vanilla absolute position encoding strictly restricts the reasoning length of LLM.  To tackle this issue, a lot of 
	advanced position encoding schemes were proposed, such as RoPE \cite{RoPE},  ALiBi \cite{Alibi}, and the recently proposed NoPE \cite{NoPE}. However, all of them only make the model architecturally-able to	deal with long inputs rather than actually perform well on long-sequence reasoning tasks \cite{li2023long, kaddour2023challenges}. 
	
	Instead of encoding the unseen length, StreamLLM \cite{streamllm} chose to discard most context but only keep the recent tokens (tokens at the end) and sink tokens (tokens at the very beginning), ensuring that the total length of the remaining context does not exceed the LLM's window size. Such a method not only enabled the LLM to deal with longer context, but also achieved a remarkable speedup. Similar idea was also adopted by LM-Infinite \cite{Lminfinite}. However,  both StreamLLM and LM-Infinite missed a lot of long-distance dependencies, leading to deviations or even errors in text understanding.
	
	The most related work should be parallel context windows (PCW) \cite{PCW}. By modifying position embedding and attention mask, PCW alleviated the context window restriction for any off-the-shelf LLM without further training. However, it was shown that PCW is only effective for a limited extension (about three times extension of its original context window size), and the performance degrades when extending to a much longer length. Moreover, the effectiveness of PCW was only demonstrated on tasks like multi-class tasks classification and information extraction. It remains an open question whether it is suitable for more general tasks.
	
	\subsection{Extension by external memory}
	Unlike the previous approaches which keep the main architecture of the model unchanged, the methods in this category usually involve modifications to the model. Generally, this kind of method introduces the external memory to restore the information of the past context, and retrieves the relevant tokens from the memory for generation based on some search mechanism, such as KNN \cite{limiformer,Memorizing}. The main disadvantage is that these methods require additional memory overhead, and they usually need further training or fine-tuning to ensure the effectiveness. \cite{InfLLM} and \cite{munkhdalai2024leave} respectively proposed an offload mechanism and a compression mechanism to reduce memory pressure.  
	
	In this paper, we aim to propose an effective training-free inference framework which enables any LLM to reason extremely long sequence.

	\section{XL$^3$M: extra-long large language model}
	Language model, including LLM, studies a conditional probability distribution (cpd) $p(x_{t+1}|X_t)$ given string of texts $X_t=[x_1,...,x_t]$\cite{wei2023overview}. 
	In the inference phase, given an input sequence, LLM also first computes the cpd and then  generates a token according to a predetermined generation mode, such as greedy search, top-k search, top-p search, etc.	Since the length of the training data is limited within a context window size $C$, LLM only studies the cpd of the cases when $t \leq C$ during training, so it fails to produce an effective cpd  at the inference stage when the length of the input sequence is larger than $C$. In other words, the context window size $C$ can be seen as an upper limit of the LLM's capacity for a single processing. Such a limit also exists in the human's reading comprehension. The human also can hardly understand a very long context by reading it from the beginning to the end at once. In fact,  we humans almost never deal with the long context in such an one-shot way. On the contrary, given a long article and a question to be answered,  we often use the following method  to get the answer:
	\paragraph{Context segmentation and key information extraction} Segment the long context first and read it segment by segment with the question. Then quickly determine which segments are relevant to the current task, and construct a short key context by reorganizing the relevant segments. Finally, answer the question based on the short key context.

	%
	%
	
	Inspired by the human's approach to reading and understanding long texts mentioned above, we propose a novel inference framework XL$^3$M, which enables any LLM to reason extremely long sequence without any continual training or fine-tuning. The  XL$^3$M  framework  follows an important principle: the length of the sequence processed by the LLM at one time should not exceed its context window size. 
	
	\subsection{Less uncertainty implies higher accuracy}
\begin{figure}[ht]
		\centering
		\vskip 0.2in
		\includegraphics[scale=0.45]{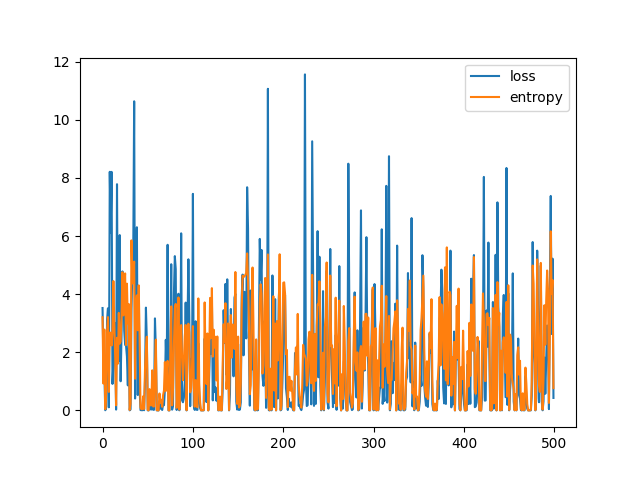}
		\caption{The relationship between the accuracy and certainty of LLM's prediction.}
		\vskip -0.2in
		\label{fig0}
	\end{figure}

	Generally, LLM is trained by optimizing the cross-entropy loss, which forces the LLM's output cpd $p(x_{t+1}|X_t)$ to gradually approach the ground-truth one-hot label vector during training. Note that the one-hot 0-1 distribution has the minimal uncertainty, so the procedure of training is also the procedure of reducing the uncertainty of LLM's prediction. Figure \ref{fig0} shows the relationship between the cross-entropy loss and the uncertainty of LLM's output cpd defined by entropy. We can see that the cross-entropy loss  and entropy value are highly positively correlated, which means that the certainty of LLM's prediction implies the accuracy to a large extent. 
	
	\subsection{Main method}
	In this subsection, we introduce the detailed procedure of XL$^3$M.  XL$^3$M follows the \textbf{context segmentation and key information extraction} path, that is, extract the relevant context first, and then reason based on the relevant context. Different from StreamLLM and other existing methods that manually discard most tokens and keep only a small part of context, XL$^3$M lets the LLM itself decide what to keep.

	%
	%
	%
	
\begin{figure}[ht]
		\vskip 0.2in
		\begin{center}
			\includegraphics[scale=0.32]{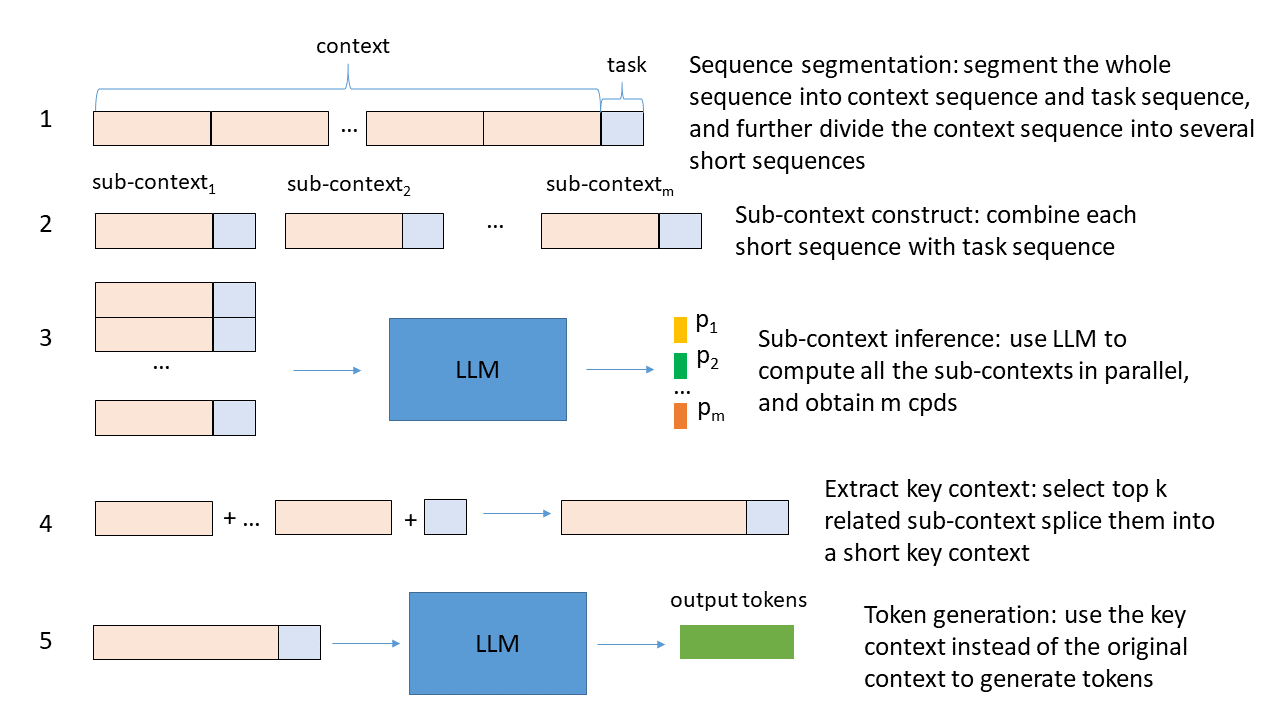}
			\caption{The main procedure of XL$^3$M.}
			\label{alpha}
		\end{center}
		\vskip -0.2in
	\end{figure}

	\paragraph{Decompose long context into short sub-contexts} Given an LLM $\Phi(\cdot|\theta)$ and an input sequence $X$ whose length is much larger than the LLM's context window size $C$, similar to \cite{PCW}, we first divide the whole sequence into a task sequence $X_t$ and a content sequence $X_c$, as shown in Figure \ref{alpha}. The task sequence is a few tokens from the end of the original input sequence, and it acts  as a ``question" to be answered. The content sequence is further segmented into $m$ short sequences $X_c^1,X_c^2,...,X_c^m$. To avoid cutting a complete sentence into two parts, we use an overlapped sliding window manner. Then by concatenating each short segment $X_c^i$ with the task sequence $X_t$, we obtain $m$ sub-contexts $X^i=[X_c^i,X_t]$, for $i=1,2,...,m$. For convenience, we call $X_c^i$ and $X_t$ the head part and end part of the sub-context $X^i$, respectively.
	
	\paragraph{Use the LLM to select relevant segments}
	We use the LLM model  $\Phi(\cdot|\theta)$ to compute the local cpd $p_i=\Phi(X^i|\theta)$ for each sub-context $X^i$.  For efficiency, all the sub-contexts can be processed in parallel. Recall that the certainty of LLM's prediction is highly correlated to the accuracy. Then we further compute the entropy for each $p_i$, namely
	\begin{equation}\label{eq:1}
	entropy(p_i) = \sum_{j=1}^{v} - p_i^j \log p_i^j,
	\end{equation}
	where $v$ is the dimension of $p_i$ and $p_i^j$ is the $j$-th element in $p_i$. A sub-context $X^i$ with small entropy value implies that the segment $X_c^i$ is relevant to the ``question" $X_t$. We select the sub-contexts $X^i$ with the top-k smallest entropy values and discard all the other noisy context. Then we construct a concise key context by splicing all the selected segments $X_c^i$ (the head part of the selected sub-contexts) as well as the task sequence  $X_t$ in chronological order.  By devising suitable segmentation and splicing strategies,  we can ensure that the length of the key context is within the training context window. 
	
	The constructed key context is used instead of the original long context to complete the inference task. Since most irrelevant content is removed and the length of the key context does not exceed the largest training length, both the out-of-domain and distraction issues are addressed.  The whole process of XL$^3$M is shown in Figure \ref{alpha}.


\section{Evaluation} \label{sec:4}
In this section, we evaluate the XL$^3$M on the comprehensive benchmark LongBench \cite{longbench} and the  widely-used long-sequence inference task ``Needle in  a Haystack'' \footnote{https://github.com/gkamradt/LLMTest$\_$NeedleInAHaystack}. All the experiments are conducted on an 8-card Huawei Ascend 910B NPU machine with 64GB memory per card.

\paragraph{Baselines} We compare the proposed XL$^3$M framework with the non-fine-tuning methods PCW \cite{PCW}, StreamLLM \cite{streamllm} and the fine-tuning models PI-7B-32k \footnote{https://huggingface.co/togethercomputer/LLaMA-2-7B-32K} and Yarn-7B-64k \footnote{https://huggingface.co/NousResearch/Yarn-Mistral-7b-64k}. PI-7B-32k is obtained by fine-tuning the  Llama2-7B-4k \footnote{https://huggingface.co/meta-llama/Llama-2-7b} model on sequences with length 32k, and Yarn-7B-64k  is obtained by  fine-tuning Mistral-7b-8k \footnote{https://huggingface.co/mistralai/Mistral-7B-v0.1} on sequences with length 64k. How to fairly compare the fine-tuning and non-fine-tuning methods itself is also an open question. Note that the model after fine-tuning actually utilizes more training data, so the model should be much more powerful. Even for  short sequence inference, the fine-tuned model still has better performance, compared with the same model before fine-tuning. For a fair comparison, the short sequence model chosen as the base model for the non-fine-tuning methods should have similar performance to the fine-tuned models on short sequence inference tasks. To achieve this goal, we construct a PI-7B-2k model by modifying the max$\_$position$\_$embeddings hyper-parameter in PI-7B-32k to 2k. The modified PI-7B-2k model has the same performance to  PI-7B-32k when the sequence length is not longer than 2k, but it is unable to address sequences longer than 2k. Namely, PI-7B-2k only inherits the capability of  PI-7B-32k in short sequence reasoning. For convenience, we use PCW-7B-2k, StreamLLM-7B-2k, XL$^3$M-7B-2k to represent the constructed PI-7B-2k models equipped with the corresponding non-fine-tuning extension methods.


\paragraph{Setup} For XL$^3$M, we use the last 128 tokens of a given sequence as the task sequence, and the rest as the content sequence. The content sequence is uniformly segmented by sliding window with overlap. The sliding window size is 512. The overlap size is 128 and for the last segment the overlap size is adjusted to ensure the consistent length of all the segments.
The initial tokens are task prompts, so we further add the initial 128 tokens to the header of each sub-context. We select the sub-contexts with the top-k (k=3) smallest entropy values as relevant sub-contexts, and use them to construct key context.  The total length of the key context is 1792, which is within 2k.	For PCW, we set the context window $n_{max}$  to be 1792 and set the number of task tokens to be 128. For StreamLLM, we use the last 1792 tokens as recent tokens and the beginning 128 tokens as sink tokens to ensure that the task prompts are included and meanwhile the total length does not exceed 2k.


\subsection{Evaluation on LongBench-E \cite{longbench}: a multitask benchmark}
LongBench is a multitask benchmark which gives a comprehensive assessment of long context understanding capabilities of LLMs. LongBench supports a subset LongBench-E, which features more evenly distributed context lengths. 
LongBench-E contains six major categories and thirteen different tasks, covering key long-text application scenarios, such as single-document QA, multi-document QA, summarization, few-shot learning, synthetic tasks, and code completion. LongBench-E includes English, Chinese, and code languages. The brief overview of LongBench-E  datasets is shown in Table \ref{statistic}. For detailed introduction of LongBench and LongBench-E, one can refer to \cite{longbench}.  In this section, we only evaluate on the English and code tasks in LongBench-E, so the MultiFieldQA-en dataset is removed, since it involves both English and Chinese.

\begin{table*}[t]
	\caption{Basic information of the dataset statistics in LongBench-E, including the data length distribution, metric, and language type.}
	\label{statistic}
	\begin{center}
		\begin{tabular}{l c c c c c}
			\toprule
			Dataset   &  0-4k  &  4-8k&  8k+ &Metric &Language \\
			\hline
			Single-Document QA &  &  & & &\\
			Qasper& 100    & 100&24&F1 &English\\
			MultiFieldQA-en &67 &70 &13&F1& English/Chinese\\
			\hline
			Multi-Document QA &&&&&\\
			HotpotQA &100 &100& 100&F1&English\\
			2WikiMultihopQA &100 & 100 &100 &F1&English\\
			\hline
			Summarization &&&&&\\
			GovReport &100 &100 &100&Rouge-L&English\\
			MultiNews &100 &100& 94&Rouge-L&English\\
			\hline
			Few-shot Learning &&&&&\\
			TREC &100 &100 &100&Accuracy (CLS)&English\\
			TriviaQA &100 &100 &100&F1&English \\
			SAMSum& 100& 100 &100&Rouge-L&English\\
			\hline
			Synthetic Task &&&&& \\
			PassageCount & 100& 100& 100&Accuracy (EM)&English\\
			PassageRetrieval-en &100 &100& 100& Accuracy (EM)&English\\
			\hline
			Code Completion&&&&&\\
			LCC &100& 100 &100& Edit Sim&Python$\verb|/|$C$\#\verb|/|$Java\\
			RepoBench-P &100& 100 &100& Edit Sim&Python$\verb|/|$Java\\
			\toprule
		\end{tabular}
	\end{center}
\end{table*}

\begin{figure*}[ht]
	\begin{center}
		\includegraphics[scale=0.3]{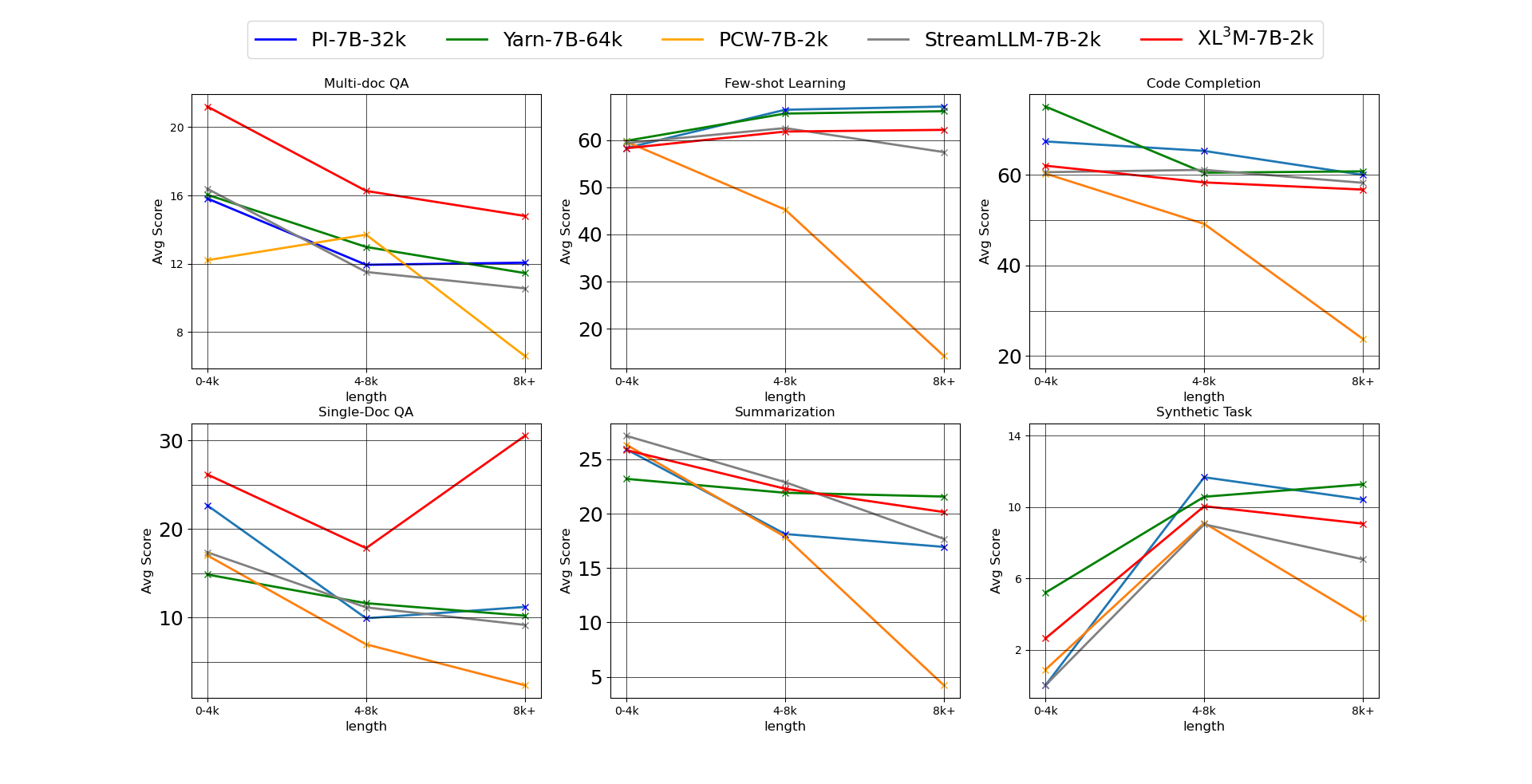}
		\caption{Average score ($\%$) under different
			context length on LongBench-E.}
		\label{fig3}
	\end{center}
\end{figure*}

We show the performance of all the compared methods on the length of 0-4k, 4-8k, 8k+, respectively. The results are shown in Figure \ref{fig3}. For each major task, we report the average score of all the datasets belonging to it. From the figure, we see that  XL$^3$M outperform all the other non-fine-tuning methods in most cases. Meanwhile, XL$^3$M  achieves comparable and sometimes even better results compared with the fine-tuning models PI-7B-32k and Yarn-7B-64k. This is mainly because our method can filter out noisy tokens and allows the model to focus on the relevant information. Note that the performance of PCW-2k drops rapidly as the length increases. This is in line with the observation shown in \cite{PCW} that PCW is only effective for a limited range of extension (about three times extension of its original context window size). Surprisingly, StreamLLM can also keep up with the performance of the other methods in some tasks, though it discards most tokens. This may be due to that the sequences in LongBench are relatively short and the answers usually appear near the end.  We will evaluate all of these methods on longer sequences  and more diverse scenarios in next subsection.

\subsection{Evaluation on ``Needle in  a Haystack'' task}
``Needle in  a Haystack'' is recently a widely-used task for testing the in-context retrieval ability of long context LLMs. The main procedure of ``Needle in  a Haystack'' is: 1. Place a random fact or statement (the ``needle'')  somewhere in a long context (the ``Haystack''); 2. Ask the model to retrieve this statement. 

\begin{figure*}[ht]
	\includegraphics[scale=0.56]{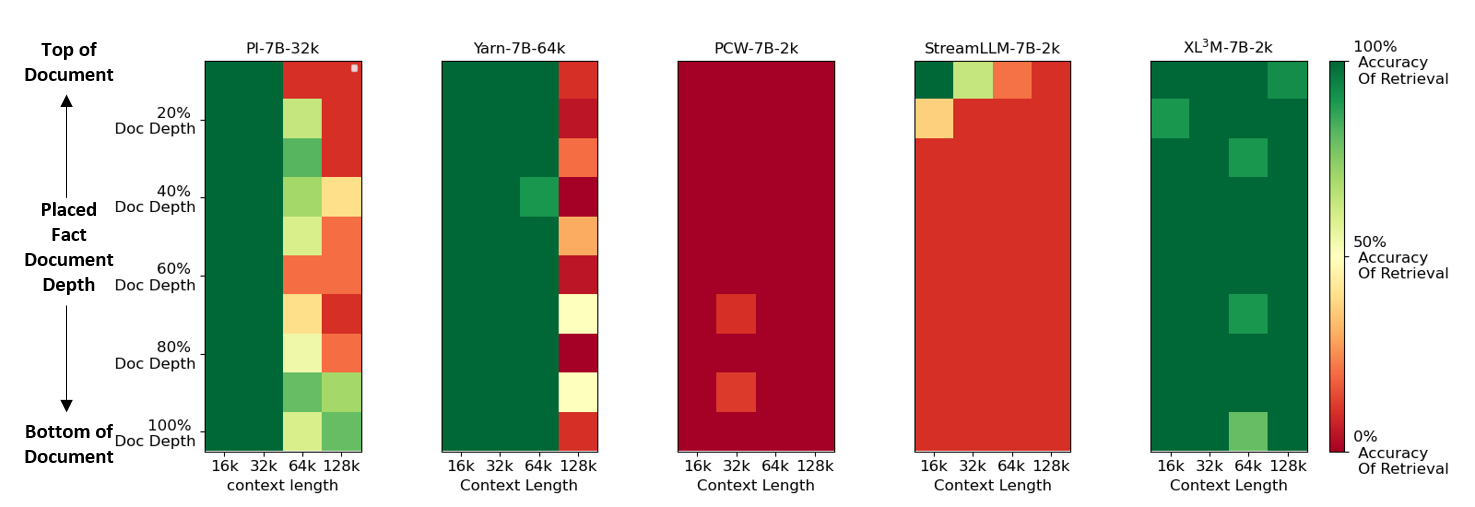}
	\caption{Pressure test on ``Needle in  a Haystack".  The test was run at 4 different lengths (16k $\to$ 128k) and 10 different ranges of document depth (buttom $\to$ top). Each result is average by 10 independent runs.}
	\label{needle}
\end{figure*}

We construct different lengths of contexts, varying from 16k to 128k, to measure the performance of all the above methods. Figure \ref{needle} shows the recall scores of all the compared methods with the ``needle'' placed at ten different ranges of depth. For each range of depth, the result is averaged by ten independent runs with the  ``needle'' randomly placed in the corresponding range each time. We can see that XL$^3$M exhibits strong performance for all the lengths,  while the PI-7B-32k and Yarn-7B-64k models only perform well when the length is within their fine-tuning context window size, and their performance  degrades rapidly  when the length exceeds the fine-tuning length. For PCW, note that the length considered in this task is 8 to 64 times larger than the context window size of PCW-7B-2k, which is beyond the effective extension range of PCW, so  it  can hardly retrieval the right answers.  StreamLLM does not perform well in this task either. Only when the ``needle'' is placed right at the end of the sequence (shallow depth) is it possible to capture the relevant information. This is inline with its mechanism of mainly keeping the recent tokens (tokens near the end) and a few initial prompt tokens. 

\begin{figure*}[ht]
	\vskip -0.2in
	\begin{center}
		\includegraphics[scale=0.31]{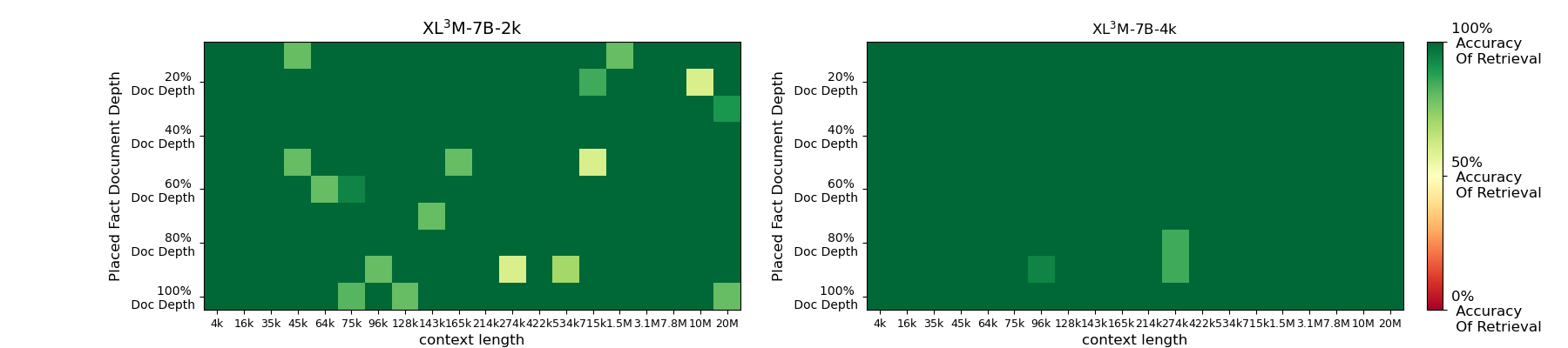}
		\caption{Pressure test on ``Needle in  a Haystack" over a larger range of lengths. Left: recall accuracy of XL$^3$M-7B-2k. Right: recall accuracy of XL$^3$M-7B-4k. }
		\label{needle2}
	\end{center}
	\vskip -0.2in
\end{figure*}

Figure \ref{needle2} further reports the performance of the proposed XL$^3$M framework over a larger range of length.  Meanwhile we investigate the impact of the context window size of the original base model on the performance of the proposed method. Using a similar manner, we construct a PI-7B-4k model by modifying the  max$\_$position$\_$embeddings hyper-parameter in PI-7B-32k to 4k. Compared with previously constructed PI-7B-2k, the only difference is that PI-7B-4k has a larger context window size. We use XL$^3$M-7B-4k to represent the  PI-7B-4k model equipped  with  XL$^3$M. The sliding window size is enlarged to 1024. All the other settings remain unchanged, so the total length of the key context is 3328, less than 4k. From the figure, we see that, XL$^3$M-7B-2k performs well in most lengths and depths, but in some cases, it does not get the $100\%$ accuracy, while XL$^3$M-7B-4k can almost retrieve the right answers for all the cases. Though the proposed XL$^3$M framework can enable any LLM to reason long sequences, the base model with a larger context window size allows a larger sliding window,  which generally contributes to achieving better performance. Moreover, the XL$^3$M framework is very memory efficient. The sequence length can be up to 20M or even larger with only 8 NPU cards.

\begin{figure}[ht]
	\vskip 0.2in
	\begin{center}
		\includegraphics[scale=0.25]{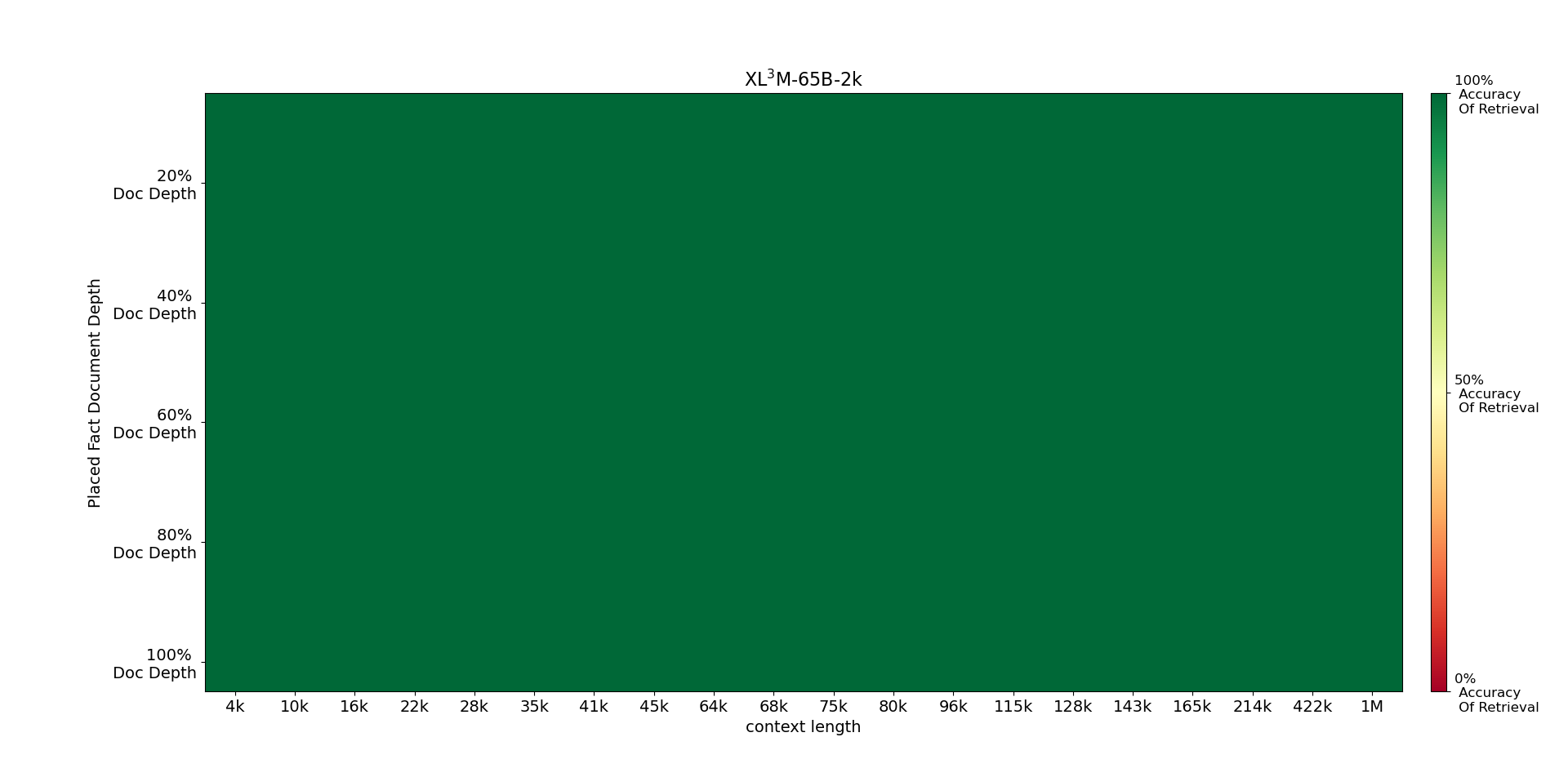}
		\caption{Test the performance of the proposed methods on Llama-65B-2k. Our methods achieve $100\%$ recalls on the  ``Needle in  a Haystack'' task for all the cases.}
		\label{llama65B}
	\end{center}
	\vskip -0.2in
\end{figure}

We also test our  XL$^3$M framework using a much larger-scale model Llama-65B  \footnote{https://huggingface.co/huggyllama/llama-65b}. The pre-training contex window size  of Llama-65B is 2k, so we follow the settings of XL$^3$M-7B-2k to pre-process the input context. We use XL$^3$M-65B-2k to represent the Llama-65B model applied with XL$^3$M. The results are shown in Figure \ref{llama65B}. We see that XL$^3$M-65B-2k  can $100 \%$ recall the answer for all the cases. This is in line with our expectation. Since Llama-65B is much more powerful than Llama2-7B, it deserves to achieve better performance. 

\subsection{Evaluation on time efficiency}
We compare the time efficiency of the proposed  XL$^3$M framework with the baselines, in terms of both prefill time (time consumption for generating the first token) and decoding time (time consumption for generating all the tokens except the first token). We evaluate all the methods on a 128k long ``Needle in  a Haystack'' task, and we set the decoding length to be 128. The compared results are shown in Table \ref{time}. The StreamLLM and XL$^3$M methods are much more time efficient than the other competitors. However, StreamLLM will introduce severe precision loss, while XL$^3$M can ensure both efficiency and effectiveness. 

\begin{table*}[t]
	\caption{Time efficiency of compared methods.}
	\label{time}
	\begin{center}
		\begin{tabular}{l c c c}
			\toprule
			Models   &  Total time (s) &  Prefill time (s)&  Decoding time (s)\\
			\hline
			PI-7B-32k &118.3&	84.2&	34.1\\
			Yarn-7B-64k&132.7 &	82.7&	50.0 \\
			PCW-7B-2k&51.2&	17.8&	33.4 \\
			StreamLLM-7B-2k &17.4&	4.3&	13.1 \\
			XL$^3$M-7B-2k&35.3&	23.1&12.2\\
			\toprule
		\end{tabular}
	\end{center}
\end{table*}

%
%

\section{Conclusion}
We empirically found that the accuracy of the LLM's prediction is highly correlated to its certainty measured by entropy. Based on this, we proposed a novel inference framework  XL$^3$M, which enables any LLM to  break the length limit without any continual training or fine-tuning. In the XL$^3$M framework, any input long context will be decomposed into multiple short sub-contexts containing a common ``question'' which is a few tokens from the end of the original input context. XL$^3$M provides a method to extract the sub-contexts relevant to the current task and discard most irrelevant context. Then a concise key context is constructed by splicing the relevant sub-contexts in chronological order. The constructed key context is further used instead of the original context to complete the inference task.  Experimental results demonstrated the effectiveness and efficiency of  XL$^3$M. We showed that equipped with XL$^3$M framework, a Llama2-7B model is able to reason 20M long sequences on an 8-card Huawei Ascend 910B NPU machine with 64GB memory per card.

\paragraph{Limitations of the proposed framework} The XL$^3$M framework assumes that  only a small portion of the original context is relevant to the given task or question, namely the length of the key context is less than the training context window size. Hence,  when the LLM's training context window is very small and relevant context contains a significant number of tokens, some key tokens has to be discarded. Moreover, when the relevant content is widely distributed in different part of the original context, it is difficult to capture all of the key context by only selecting a few segments. For these cases, XL$^3$M may not achieve satisfactory performance.

\bibliography{example_paper}
\bibliographystyle{icml2023}

\appendix

\end{document}